\documentclass[conference]{IEEEtran}

\IEEEoverridecommandlockouts

\usepackage{graphicx} 
\usepackage{amssymb} % Therefore symbol
\usepackage{amsmath} 
\usepackage{subcaption}
\usepackage{multirow}
\usepackage{adjustbox}
\usepackage{placeins}
\usepackage{marvosym}
\usepackage{makecell} 
\usepackage{arydshln}      % for dashed lines
\usepackage{booktabs}
\usepackage{stfloats}
\usepackage{caption}    % for \captionof
\usepackage{cuted}
\usepackage{hyperref}

\usepackage[inline]{enumitem}

\usepackage{xspace}
\usepackage[textwidth=1.4cm,textsize=scriptsize]{todonotes}
\definecolor{colorgb}{RGB}{255, 150, 0}

\title{\LARGE SPADE: Sparsity Adaptive Depth Estimator for Zero-Shot, Real-Time, Monocular Depth Estimation in Underwater Environments\\
\thanks{This work has been submitted to IEEE Journal of Oceanic Engineering Special Issue of AUV 2026}
\vspace{-5pt}
}

\author{
\IEEEauthorblockN{Hongjie Zhang, Gideon Billings, Stefan B. Williams} % <-this % stops a space
\IEEEauthorblockA{\textit{Australian Centre for Robotics}\\ \textit{University of Sydney}\\ Sydney, NSW 2006, Australia\\ Email: {hongjie.zhang, gideon.billings, stefan.williams}@sydney.edu.au}
\vspace{-25pt}
}

\begin{document}

\maketitle

\begin{abstract}
Underwater infrastructure requires frequent inspection and maintenance due to harsh marine conditions. Current reliance on human divers or remotely operated vehicles is limited by perceptual and operational challenges, especially around complex structures or in turbid water. Enhancing the spatial awareness of underwater vehicles is key to reducing piloting risks and enabling greater autonomy. To address these challenges, we present SPADE: SParsity Adaptive Depth Estimator, a monocular depth estimation pipeline that combines pre-trained relative depth estimator with sparse depth priors to produce dense, metric scale depth maps. Our two-stage approach first scales the relative depth map with the sparse depth points, then refines the final metric prediction with our proposed Cascade Conv-Deformable Transformer blocks. Our approach achieves improved accuracy and generalisation over state-of-the-art baselines and runs efficiently at over 15 FPS on embedded hardware, promising to support practical underwater inspection and intervention. 
\end{abstract}

\vspace{-10pt}

\section{Introduction}
For both Remotely Operated Vehicles (ROVs) and Autonomous Underwater Vehicles (AUVs), situational awareness is critical for the execution of complex intervention tasks, such as ship hull inspection, infrastructure maintenance, cleaning, or sample collection. While remote ROV pilots often operate using monocular RGB camera streams, online 3D reconstruction of the vehicle workspace has been shown to reduce risks and increase operational tempo for piloted control~\cite{phung2024shared}. For autonomous systems, online 3D scene reconstruction is critical for safely and accurately executing intervention tasks.

Active range sensing techniques, including LiDAR, Time-of-Flight (ToF), and RGB-D cameras are extensively used in terrestrial applications for depth estimation and scene reconstruction. However, the effectiveness of these active sensors is reduced underwater, due to inherent attenuation and back-scattering effects~\cite{FLSea}. Acoustic sensing, particularly forward-looking 2D imaging sonar, serves as an alternative underwater range sensor but provides only a top-down, 2D view, introducing elevation ambiguity. Recent developments in 3D imaging sonar show potential for operating similar to LiDAR in providing sparse point clouds. However, sonar sensors in general are limited in resolution and signal to noise ratio compared to their optical sensor counterparts. While previous studies \cite{mono_single_beam, camera_sonar_slam} explored dense depth perception using imaging sonar, effectively utilising this sensing modality in online scene reconstruction remains an ongoing research challenge.

RGB cameras with artificial lighting systems are widely employed underwater due to their simplicity and low cost relative to active sensors. Stereo camera systems provide strong feature matches and can efficiently estimate depth through stereo correlation, but their performance underwater is frequently compromised by poor textures, strong lighting effects, and low illumination. Additionally, the quality of stereo depth is sensitive to the chosen stereo baseline, which is subject to hardware and space constraints. Monocular cameras offer an alternative solution, being low-cost, compact, and easily integrated into various underwater robotic platforms, but they inherently lack scale information. Recent advancements in deep learning have driven substantial progress in monocular metric depth estimation~\cite{midas3.1}~\cite{DepthAnything}, which motivates this work.

\begin{figure*}
  \centering
  \includegraphics[width=0.85\linewidth]{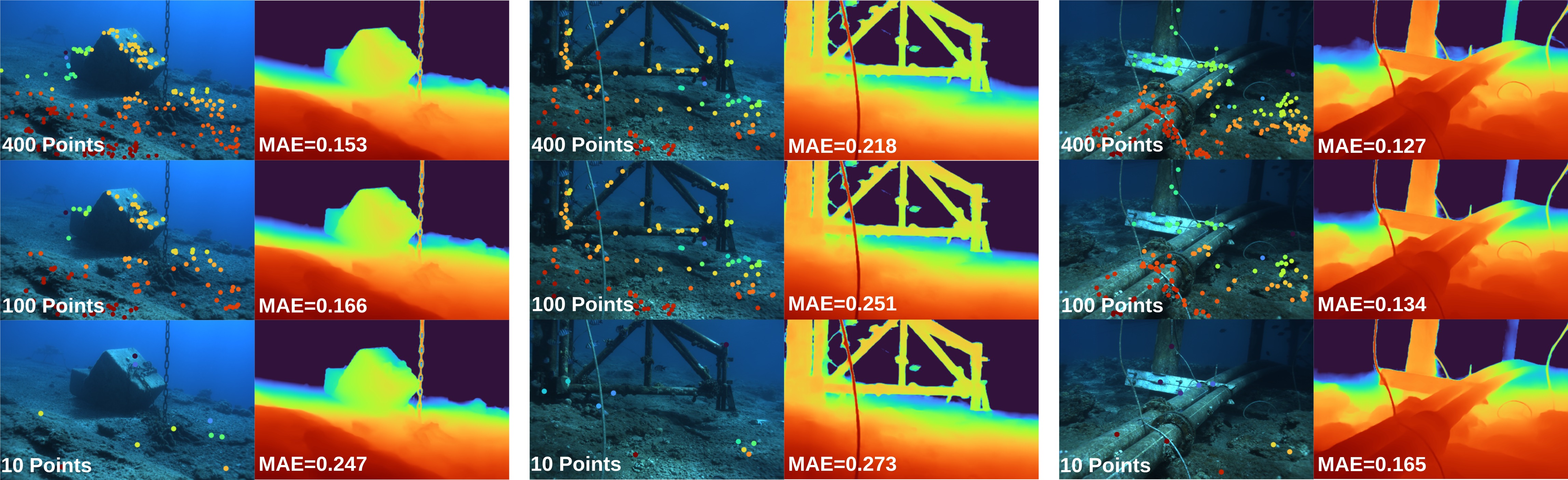}
  \caption{\small Our proposed monocular depth estimation pipeline takes RGB images and sparse depth points as input and predicts dense metric depth maps. It shows great generalisation on underwater data and strong robustness against varying sparsity of the depth points.}
  \vspace{-10pt}
  \label{fig:sparsity_prediction}
\end{figure*}

To address the challenges of underwater spatial perception, we propose \textbf{SPADE}: \textbf{SP}arsity \textbf{A}daptive \textbf{D}epth \textbf{E}stimator, a robust, compact, monocular depth estimation pipeline for underwater robotic applications. As illustrated in Fig.~\ref{fig:sparsity_prediction}, our approach demonstrates strong zero-shot generalization in underwater scenes and remains robust to varying levels of depth prior sparsity. The contributions in this paper include:
\begin{itemize}
\item The development of a monocular underwater depth estimation pipeline that integrates a relative depth estimator with sparse depth priors to produce dense metric depth. 
\item The introduction of the Cascade Conv-Deformable Transformer (CCDT) block, which combines convolutional and transformer modules for improved scale refinement.
\item The implementation of an efficient, real-time system achieving over 15 FPS on Jetson Orin NX.
\item The demonstration of improved performance on diverse underwater datasets relative to other state-of-the-art methods, especially when sparse depth measurements are limited. Our evaluation is available at: \url{https:/github.com/Jayzhang2333/Sparsity-Adaptive-Depth-Estimator}.
\end{itemize}

\section{Related Work}
\subsection{Monocular Depth Estimation}
Supervised deep learning approaches have effectively learned to estimate relative depth maps by leveraging colour and geometric cues from images \cite{DepthAnything, midas3.1, marigold}. Current state-of-the-art monocular relative depth estimation models utilise transformer-based architectures, such as Vision Transformers (ViTs), coupled with specialized loss functions such as scale-invariant and scale-shift-invariant losses \cite{midas3.1}. These models predict affine-invariant depth maps, which can be transformed into metric measurements if appropriate scaling information is available. These state-of-the-art models have been shown to generalise well across diverse scenes and varying environmental conditions through training on large scale mixed datasets \cite{midas3.1}. More recently, a method based on stable diffusion also leverages latent encoders pre-trained on large-scale datasets to enhance generalisation performance \cite{marigold}.

While supervised relative depth estimation methods have achieved impressive zero-shot performance on underwater imagery, where training data is scarce~\cite{Atlantis}, naive metric scaling results in limited accuracy. Model performance in unseen environments can be substantially improved through transfer learning techniques that fine-tune on small, specialised datasets. However, this approach can lead to overfitting and reduced generalisation of the relative depth model~\cite{videpth}, particularly in underwater environments where variability is high. Nonetheless, these state-of-the-art relative models still encode strong knowledge of the general 3D scene structure, which can be leveraged for fine-grained depth prediction.

\subsection{Depth Estimation with Sparse Measurements}
To resolve scale ambiguity, sparse depth priors from LiDAR measurements, stereo matching or visual-inertial odometry, are often projected into the camera frame and fused with RGB images to predict dense metric depth. Early fusion combines RGB images and sparse depth maps at the input, enabling the encoder to extract integrated features. State-of-the-art methods use both transformer and convolutional blocks to capture spatial detail and global context \cite{completionformer}. The approach proposed in \cite{SparseDC} employs gated convolutions to transform the combined RGB and sparse depth input into rich feature representations. Alternatively, some approaches fuse sparse depth maps during the decoder stage by integrating enriched metric features into decoder feature maps at each stage \cite{SpAgNet, FlexibleDC}. Both fusion strategies frequently use refinement modules such as Spatial Propagation Networks (SPNs) \cite{NLSPN} before the final output depth, which iteratively improve depth estimates based on pixel affinity maps from specialized decoder outputs.

Although these approaches achieve good accuracy, they often overfit to specific scenarios, such as depth ranges, environments, or sensor-specific biases related to the distribution and range of sparse depth points \cite{VirtualStereo}. Recent work has improved generalisation  across range and environments by combining sparse depth measurements with vision backbones, relative depth estimators, or foundation models pretrained on large datasets \cite{uwdepth, videpth, DepthPrompting, MarigoldDC}. The method most similar to ours is VI-Depth \cite{videpth}, which uses a two-stage process: a pretrained MiDaS network first predicts an affine-invariant depth map, which is aligned to metric scale using sparse visual-inertial odometry (VIO) measurements, followed by a second network that refines the metric depth map with per-pixel scale corrections. While VI-Depth demonstrates strong generalisation across scenes, its performance declines with fewer or clustered sparse depth points. Similar approaches \cite{uwdepth, DepthPrompting} also utilise pretrained relative depth estimation networks to produce initial depth priors and incorporate multi-scale, depth-aware features from various decoder stages. Our work builds on VI-Depth, with a focus on improving accuracy, and robustness under challenging sparse depth conditions.

\section{Method}
\begin{figure*}[htbp]
  \centering
  \includegraphics[width=0.85\textwidth]{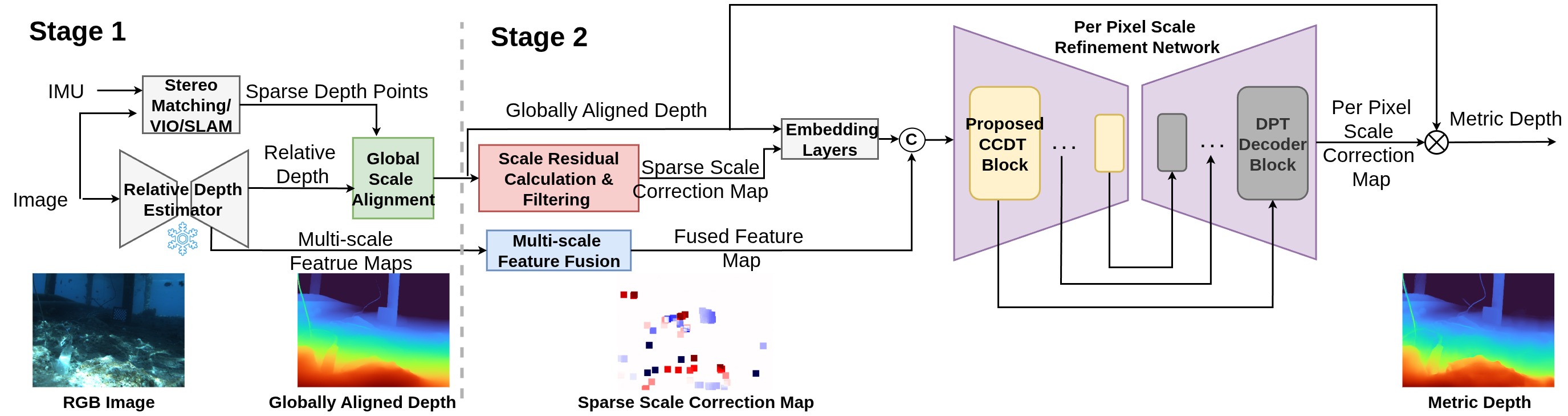} 
  \caption{\small The SPADE pipeline consists of two stages. Stage 1 generates sparse metric depth points using SLAM or single-shot stereo matching, and a monocular depth estimator predicts a relative depth map from an RGB image. This map is then aligned to the sparse measurements to obtain metric scale. In Stage 2, a scale refinement network built with our Cascade Conv-Deformable Transformer (CCDT) block, uses the sparse scale correction map, aligned depth map, and fused multi-scale features from relative depth estimator's encoder to predict a per-pixel scale correction map, which is then multiplied with the aligned depth map to produce the final dense depth estimate.}
  \vspace{-10pt}
  \label{fig:overall pipeline diagram}
\end{figure*}

Fig.~\ref{fig:overall pipeline diagram} shows an overview of the SPADE depth estimation pipeline. The pipeline is inspired by~\cite{videpth}, and contains two main stages: (1) sparse points generation, monocular relative depth estimation and global alignment, and (2) per-pixel scale refinement to produce the final metric depth predictions. 

\subsection{Sparse Depth Generation}
Depending on the specific system setup, SPADE can use monocular camera and IMU via visual-inertial SLAM to generate accurate sparse depth, or single-shot stereo matching and triangulation to achieve single-shot depth estimation if stereo data is available. In both cases, we employ SIFT feature descriptors, as work in~\cite{CUDASiftSLAM} has shown their robustness in underwater conditions. During evaluation, we use either method depending on the availability of each dataset. Although our approach can leverage strong stereo constraints via epipolar filtering, the core network operates on a monocular image and sparse depth map, ensuring lower computational cost compared to stereo networks with parallel branches.

\subsection{Monocular Depth Estimation and Global Alignment}
The monocular depth estimation network used in our pipeline is DepthAnythingV2 \cite{DepthAnything}, a pretrained state-of-the-art relative depth estimator. This model takes an RGB image as input and outputs an estimated affine-invariant depth map. By training with large scale data, this relative depth estimator achieved strong generalisation performance \cite{DepthAnything}. In this work, the lightweight version DepthAnythingV2 Small is used to achieve real-time performance.

Then the affine‐invariant prediction from DepthAnythingV2 is converted into metric scale by adopting the global alignment procedure from \cite{videpth}. Specifically, the affine-invariant depth $z$, which represents unitless depth values in inverse depth space can be converted into globally aligned metric depth $\tilde{z}$ by applying a global scale $s$ and shift $t$ as~\eqref{eq:affine_transform}. The optimal scale $s$ and shift $t$ can be found by minimising a least squares error, as defined in~\eqref{eq:scale_shift_least_square}, where the sparse depth measurements, denoted by $d$, are converted into inverse depth space as $v = \frac{1}{d}$ and $N$ is the total number of sparse depth points.
\begin{gather}
  \tilde{z} = s\,z + t
    \label{eq:affine_transform}\\
  (s, t) = \arg\min_{s,t}\sum_{i=1}^{N}\bigl((s\,z_i + t) - v_i\bigr)^2
    \label{eq:scale_shift_least_square}
\end{gather}

In practice, we found that occasionally when sparse depth points are clustered, jointly optimising scale and shift can overfit the shift term, resulting in a negative scale that inverts depth relations and distorts scene structure. When such cases happen, we convert to scale-only globally aligned metric depth $\tilde{z}_{scale}$ by applying only a global scale term to $z$ as shown as~\eqref{eq:affine_transform_scale_only}. The optimal scale is calculated by minimising~\eqref{eq:scale_least_square}. Applying only this scale can still align affine-invariant depth with the metric depth magnitude, though it may require the refinement network to make larger corrections.
\begin{gather}
\tilde{z}_{scale} = s\,z \label{eq:affine_transform_scale_only}\\
s = \arg\min_{s} \sum_{i=1}^{N} \bigl(s\,z_i - v_i\bigr)^2
\label{eq:scale_least_square}
\end{gather}

\subsection{Per-pixel Scale Refinement Network}
\subsubsection{Network Overview}
The globally aligned depth map $\tilde{z}$ offers only a coarse metric depth estimate and is often inaccurate across different image regions. To correct the prediction at pixel level, we use a refinement network to predict a per-pixel scale correction map $\hat{\epsilon}$. Multiplying this map with the globally aligned depth map yields the final refined metric depth $\hat{z}$ as shown in~\eqref{eq:apply scale correction}, which is in inverse depth space. Compared with directly regressing absolute depth, this strategy improves generalisation across scenes with widely varying depth ranges.
\begin{equation}
  \hat{z} =  \tilde{z} \odot \hat{\epsilon}
    \label{eq:apply scale correction}
 \end{equation}

Given the sparse inverse depth points $v$ and the globally aligned depth map $\tilde{z}$, a sparse scale correction map $\epsilon$ can be computed as~\eqref{eq:per_pixel_scale_corr}. Combining with the other two inputs: the globally aligned depth map $\tilde{z}$, and the fused multi-scale feature map from DepthAnything’s DINOv2 encoder, the refinement network predicts a dense per-pixel scale correction map.
\begin{equation}
  \epsilon = \frac{v}{\tilde{z}}
    \label{eq:per_pixel_scale_corr}
 \end{equation}

The refinement network follows the U-Net \cite{U-Net} structure. The sparse scale correction map and the globally aligned depth map first pass through each one's embedding layer, their embeddings are then concatenated with the DINOv2 features to form the encoder input. The encoder, built from our Cascade Conv–Deformable Transformer blocks, downsamples the spatial resolution to \(\tfrac14\), \(\tfrac18\), \(\tfrac1{16}\), and \(\tfrac1{32}\) of the input to extract features while propagating the sparse known scale correction factors to unknown regions. The decoder comprises four DPT decoder blocks~\cite{DPT} that progressively upsample and fuse the encoder features with skip connections. Before each fusion, the encoder’s feature maps pass through a convolutional projection layer to ensure matching channel dimensions. The decoder outputs the per-pixel scale-correction map via an output convolution head.

\subsubsection{Upsampling Sparse Scale Correction Map} 
To avoid the network receiving an overly sparse input, we densify the sparse scale correction map using joint bilateral upsampling~\cite{joint_bilateral_upsampling} before inputting to the refinement network. This method uses the globally aligned depth \(\tilde z\) as guidance to propagate the known scale correction factors to neighbouring pixels of similar depth. The upsampled scale correction factor value at pixel \(p\) is given by~\eqref{eq:filtered_scale_residual} and the remaining zeros are replaced with a default correction factor of 1.
\begin{gather}
  \tilde{\epsilon}_p = \frac{1}{k_p} \sum_{q \in \Omega} \epsilon_q \, f(\|p - q\|) \, g(\|\tilde{z}_p - \tilde{z}_q\|),
    \label{eq:filtered_scale_residual}
\end{gather}
where $f$ is the spatial kernel, $g$ is the range kernel and $k_p$ is the normalisation factor.

\subsubsection{Multi-scale Feature Map}
To provide extra photometric and geometric cues, we iteratively fuse the four multi-scale feature maps from DepthAnything’s DINOv2 encoder into a 32-channel feature map as input to the network. At each fusion step the current feature map is upsampled, concatenated with the feature map from the next higher resolution, and passed through a \(1 \times 1\) convolution. 

\subsubsection{Cascade Conv-Deformable Transformer Encoder Block} 
\begin{figure}[bp]
  \centering
  \includegraphics[width=0.45\textwidth]{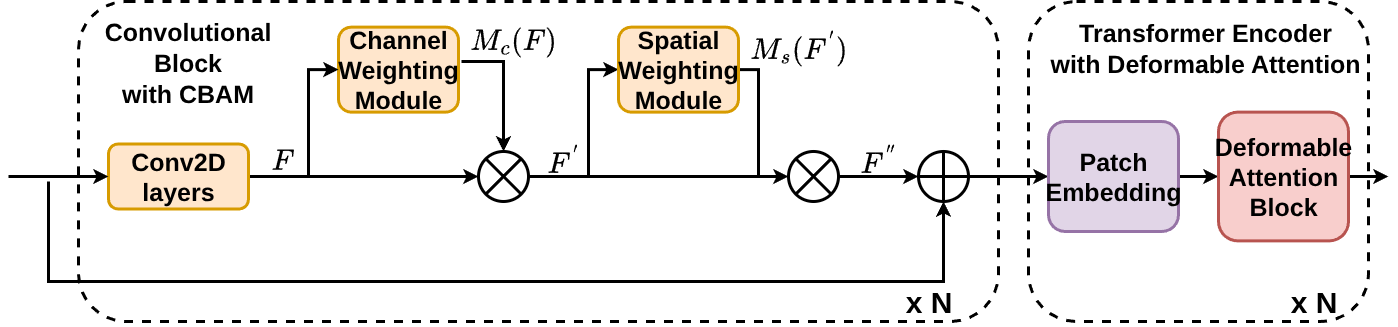} 
  \caption{\small The Cascade Conv-Deformable Transformer block has two main components, ResNet Block with the Convolutional Block Attention Module (CBAM) \cite{CBAM} and the transformer encoder with deformable attention \cite{dat++}.}
  \label{fig:encoder block}
  \vspace{-5pt}
\end{figure}

Underwater scenes range from feature-rich to almost featureless, so the captured sparse depth points can be dense and well-distributed in one frame, yet extremely scarce or clustered in the next. Since initial scale correction factors are computed from the depth points, these variations are directly reflected in the sparsity of the input scale correction map. To maintain optimal performance under both conditions, the network must (i) leverage all local scale correction cues and (ii) exploit reliable features from spatially distant scale correction hints. Inspired by \cite{completionformer}, we propose the Cascade Conv-Deformable Transformer encoder block (Fig.~\ref{fig:encoder block}), which integrates convolutional locality with transformer-based global context.

As illustrated in Fig.~\ref{fig:encoder block}, the convolutional sub-block first extracts rich local features, effectively propagating scale cues from pixels with known scale correction factor to neighbouring pixels. A transformer encoder sub-block then relates these enhanced features across the entire image, enabling the network to extract useful scale correction cues from even distant regions. Within a single CCDT block, multiple convolutional transformer sub-blocks can be stacked as needed, giving it adjustable capacity for capturing local or global features. This flexibility also enables the encoder to be designed for lightweight, real-time models or heavier, high-accuracy configurations for offline processing.

% CBAM
We improve the capability of the ResNet convolutional block to extract sparse local scale correction features by adding the Convolutional Block Attention Module (CBAM) proposed in \cite{CBAM}. As shown in Fig.~\ref{fig:encoder block}, after the convolution layers produce a feature map $\mathbf{F}\in\mathbb{R}^{C\times H\times W}$, the CBAM block generates a channel attention map $\mathbf{M}_c(\mathbf{F})$ and a spatial attention map $\mathbf{M}_{s}(\mathbf{F}^{'})$. These attention maps are multiplied sequentially to the output feature map from the convolution layers as~\eqref{eq:apply channel atention} and~\eqref{eq:apply spatial atention} to amplify the most informative feature channels and emphasise critical neighbours. 
\begin{gather}
\mathbf{F}^{'} = \mathbf{M}_c(\mathbf{F}) \odot \mathbf{F} \label{eq:apply channel atention}\\
\mathbf{F}^{''} = \mathbf{M}_s(\mathbf{F}^{'}) \odot \mathbf{F}^{'} \label{eq:apply spatial atention}
\end{gather}

\begin{figure*}[htbp]
  \centering
  \includegraphics[width=0.65\textwidth]{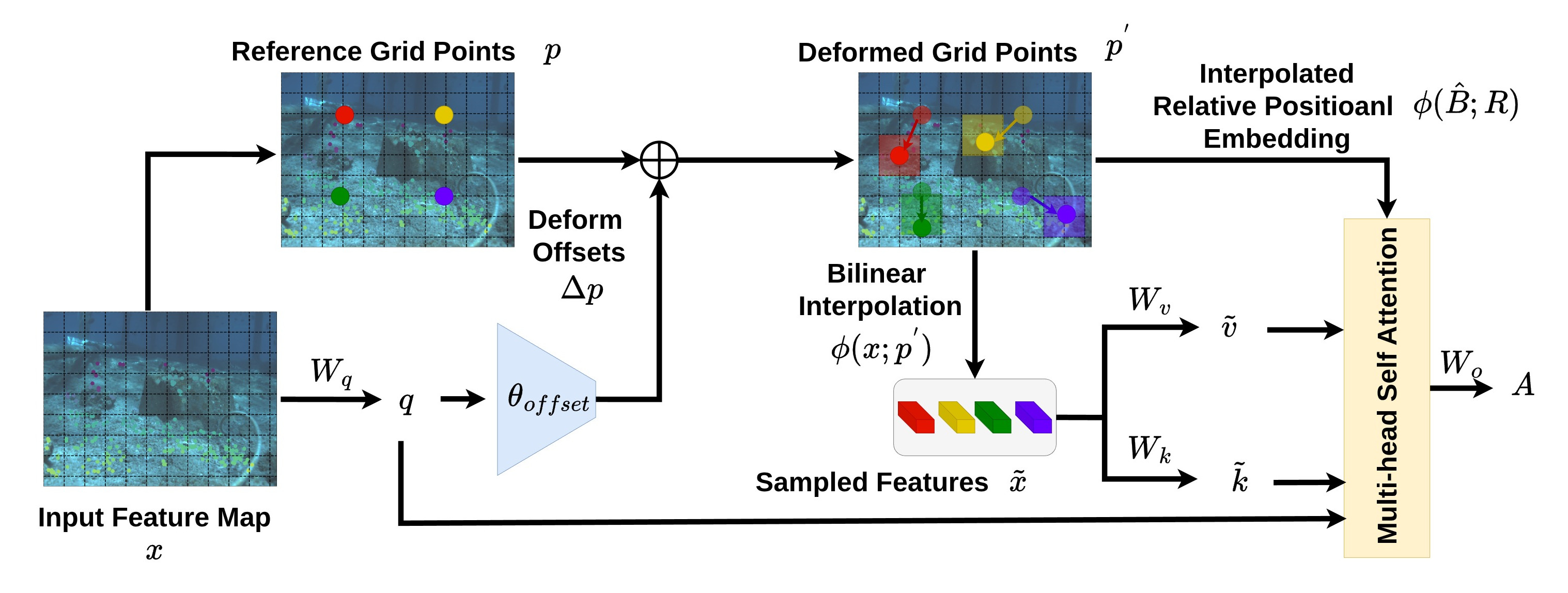} 
  \vspace{-5pt}
  \caption{\small In deformable attention operation, a reference grid $p$ is constructed over the input feature map $x$. All features are projected into queries $q$ and an offset network $\theta_{offset}$ takes in the queries and predicts spatial offsets $\Delta p$ to the grid points to shift them to more informative regions $p^{'}$. Then a subset of the features $\tilde{x}$ are sampled at the deformed grid points through bilinear interpolation. The sampled features are then projected into key-value pairs and multi head attention is calculated between the queries and the sampled key-value pairs. This diagram is an adaptation of the original DAT diagram in~\cite{dat++}.}
  \vspace{-10pt}
  \label{fig:deformale operation}
\end{figure*}

We adopt the Vision Transformer Encoder with Deformable Multi-head Attention (DAT) \cite{dat++} to efficiently extract global context from the most informative regions. As illustrated in Fig.~\ref{fig:deformale operation} and~\eqref{eq:deformable_attention}, unlike the standard self-attention in the vision transformer encoder \cite{vit}, deformable attention attends queries only to key–value pairs projected from sampled features. 

These sampled features \(\tilde{x}\) are obtained by first initialising a subsampled reference grid \(p\) over the input feature map \(x\). An offset network \(\theta_{offset}\) then takes the queries \(q\) as input and predicts offsets \(\Delta p\) for each grid node to shift them to the most informative regions: \(p' = p + \Delta p\). Since \(p'\) is continuous, the features are sampled at these locations using bilinear interpolation \(\phi(\,\cdot\,;\,\cdot\,)\). The resulting interpolated features \(\tilde{x}\) are then projected into keys $\tilde{k}$ and values $\tilde{v}$.

To preserve spatial context, DAT uses the relative positional embedding table \(\hat{B}\) \cite{Swin}, which encodes relative displacements between queries and keys. Because the deformed key positions are continuous, we also obtain their relative embeddings by applying bilinear interpolation to \(\hat{B}\) at the continuous relative displacements \(R\).

Finally, multi-head attention is calculated between queries and the sampled key-value pairs as follows:
\begin{equation}
A^{(n)} = \sigma\left(\frac{q^{(n)} \tilde{k}^{(n)\top}}{\sqrt{d}} + \phi(\hat{B}; R)\right)\tilde{v}^{(n)}, \label{eq:deformable_attention}
\end{equation}
where $A^{n}$ is the $n^{\text{th}}$ head output of $N$ total heads and $d = C/N$ is the dimension of each head. The attention outputs from multiple heads are concatenated and projected to the final output tensor $A$.
\begin{equation}
A = [A^{(1)}, \dots, A^{(n)}]W_0
\end{equation}

To our knowledge, this is the first time the deformable attention transformer has been applied to a depth estimation or depth completion task. Compared to efficient vision transformer alternatives with handcrafted, data-agnostic attention (e.g., Swin Transformer (SWIN) \cite{Swin} or Pyramid Vision Transformer (PVT) \cite{PVT}), deformable attention adaptively focuses on the more informative key–value pairs for each input, reducing the risk of important information from being missed in fixed attention patterns~\cite{dat++}. This adaptability is well suited to underwater environments, where the distribution of sparse depth points varies significantly from frame to frame. Additionally, by incorporating the globally aligned depth map and the DINOv2 feature maps, the transformer encoder learns query–key affinities not only from scale-correction cues but also from geometric and photometric features, enhancing its ability to identify and leverage relevant distant regions.

\subsection{Implementation Details}
Given our aim for a real-time depth estimation pipeline, we designed a lightweight refinement network whose number of convolutional and transformer sub-blocks per stage’s CCDT encoder is detailed in Table~\ref{tab:refinement_net_summary}. Combined with the DepthAnythingV2 Small backbone (DA V2 Small), our model remains compact. Leveraging stereo imagery with single-shot feature extraction and matching to generate sparse depth points, alongside TensorRT FP16 optimisation of the network model, our full depth estimation pipeline achieves $\geq 15$ FPS on an NVIDIA Jetson Orin NX.
\begin{table}[htbp]
  \centering
  \caption{\small Refinement network implementation details.}
  \label{tab:refinement_net_summary}
  \scriptsize
  \begin{tabular}{lccc}
    \toprule
    Component        & \#Conv Blocks & \#Trans Blocks & Params (M) \\
    \midrule
    Refinement Net   & [1,1,2,2]     & [1,1,2,2]      & 29.36      \\
    DA V2 Small      & --            & --             & 24.80      \\
    \midrule
    \textbf{Total}   & --            & --             & \textbf{54.16} \\
    \bottomrule
  \end{tabular}
  \vspace{-12pt}
\end{table}

\subsection{Loss Function} 
We use the Root Mean Square Error (RMSE) loss to supervise precise pixel-level depth accuracy:
\begin{equation}
    \mathcal{L}_{\mathrm{RMSE}}(\hat z, z^*) 
    = \sqrt{\frac{1}{N}\sum_{i=1}^N (\hat z_i - z^*_i)^2},
\label{eq:rmse}
\end{equation}
the Scale-Invariant Log Loss (SiLog) to maintain relative depth relationships within the scene:
\begin{equation}
\mathcal{L}_{\mathrm{SiLog}}(\hat z, z^*) 
= \beta \sqrt{\frac{1}{N}\sum_{i=1}^N g_i^2 
   - \frac{\lambda}{N^2}\left(\sum_{i=1}^N g_i\right)^2},
\label{eq:silog}
\end{equation}
where $g_i = \log \hat z_i - \log z^*_i$, $\lambda = 0.85$, and $\beta = 10$.  
We also use the multi-scale gradient loss to preserve sharp edges in the predicted depth maps:
\begin{equation}
\mathcal{L}_{\mathrm{grad}}(\hat z, z^*) 
= \frac{1}{K}\sum_{k=1}^{K}\frac{1}{M}\sum_{i=1}^{M}
\bigl(|\nabla_x R_i^k| + |\nabla_y R_i^k|\bigr),
\label{eq:grad}
\end{equation}
where $R_i^k = z_i^{*k} - \hat z_i^k$ and $K = 3$.  

These losses are computed in inverse-depth space to prioritise close-range accuracy, which is critical for underwater intervention tasks. We denote the predicted inverse depth as $\hat{z}$ and the ground truth inverse depth as $z^*$. The overall loss is the weighted sum of the three components, as given in \eqref{eq:overall_loss}.
\begin{equation}
    \mathcal{L} = \mathcal{L}_{\mathrm{RMSE}} + \mathcal{L}_{\mathrm{SiLog}} + 0.5 \, \mathcal{L}_{\mathrm{grad}}
\label{eq:overall_loss}
\end{equation}

\section{Experiments}
Existing real-world underwater datasets with ground truth metric depth maps are scarce. The ground truth depth maps are typically generated from stereo matching or Structure-from-Motion (SfM), which often yields incomplete depth maps with limited accuracy. To avoid over-fitting on a particular underwater scene or camera settings, we trained our model purely on the TartanAir \cite{TartanAir} dataset, we then evaluate the zero-shot generalisation of our method to a diverse set of real-world underwater scenarios. Table~\ref{tab:dataset_settings} summarises the datasets used in our work. 

TartanAir \cite{TartanAir} is a synthetic dataset with dense, per-pixel metric depth maps. To generate sparse depth measurements, the front end of VINS-Mono SLAM~\cite{VINS-MONO} is run on the TartanAir dataset to extract the pixel locations of tracked visual feature points, and the depth values at these location are then extracted from the ground truth depth maps. On average, we extract 250 sparse depth points per image. During training, we randomly sample 90\% of these points as a form of data augmentation, which also helps prevent overfitting to any particular sparse point pattern.
\begin{table}[htbp]
  \centering
  % locally reduce font and tweak spacing
  % \footnotesize
  \setlength{\tabcolsep}{4pt}            % reduce column padding
  \renewcommand{\arraystretch}{1.2}      % increase row height

  \caption{\small Summary of dataset settings}
  \label{tab:dataset_settings}
  \begin{adjustbox}{width=\columnwidth}
    \begin{tabular}{l p{2cm} c c c}
      \hline
      \textbf{Dataset}    
        & \textbf{Scenes}                      
        & \textbf{\# Images} 
        & \textbf{GT Depth Source}              
        & \textbf{Usage} \\ 
      \hline
      TartanAir \cite{TartanAir} 
        & \makecell[l]{16 different scenes\\(indoor, outdoor,\\underwater)} 
        & 120K+              
        & \makecell[l]{Dense synthetic\\ depth map  }                        
        & Training       \\
      \hline
      FLSea VI \cite{FLSea}      
        & \makecell[l]{Underwater\\forward-looking\\monocular}              
        & 3270       
        & \makecell[l]{Monocular \\Structure-from-Motion}                       
        & Testing        \\
     \hline
      Lizard Island \cite{CUDASiftSLAM}              
        & \makecell[l]{Underwater\\downward-looking\\stereo}                
        & 2216       
        & \makecell[l]{Stereo \\ Structure-from-Motion}                   
        & Testing        \\
     \hline
      Water Tank                 
        & \makecell[l]{Handcrafted object\\in water tank}                  
        & 34                 
        & \makecell[l]{Rendered depth map\\ with camera pose \\ and 3D model}  
        & Testing        \\
      \hline
    \end{tabular}
  \end{adjustbox}
  \vspace{-10pt}
\end{table}

\subsection{Training Setup}
To leverage the generalisation performance of the pre-trained monocular depth estimator, we freeze the DepthAnything model during training. We train for 10 epochs using the AdamW optimiser with a batch size of 16. The learning rate is set to $2\times10^{-4}$ for the first 6 epochs, then decayed to $5\times10^{-5}$ for the remaining 4 epochs. Input images and sparse depth maps are resized to a resolution of $336\times448$.

\subsection{Evaluation Metrics}
To evaluate our pipeline, we compute Root Mean Square Error (RMSE), Mean Absolute Error (MAE), and Absolute Relative Error (AbsRel) between the predicted depths $\hat d$ and the ground truth $d^*$ to quantify per‐pixel accuracy. We also report the Scale‐Invariant Log error ~\eqref{eq:silog} to assess how well the model preserves the scene’s relative depth relationships and overall structure:

\begin{equation}
\mathrm{SILog} 
= \sqrt{\frac{1}{N}\sum_{i=1}^{N}\left(\log \hat d_i - \log d_i + \alpha \right)^2},
\label{eq:silog}
\end{equation}
where $\alpha = \tfrac{1}{N}\sum_{i=1}^{N}\bigl(\log d_i - \log \hat d_i \bigr)$.

\subsection{Test with FLSea Dataset}
The FLSea VI dataset \cite{FLSea} is used to assess our pipeline’s performance on forward‐looking, close and long‐range underwater scenes. The ground truth depth maps are generated via SfM. For evaluation, we selected over 3,000 images including images from a close-range canyon and an open-water area with artificial structures. Although this dataset includes both monocular images and IMU data, each sequence lacks proper initialisation, resulting in poor visual–inertial SLAM outputs. To mimic sparse depth generation using SLAM methods, we detect and track SIFT feature points across each image sequence and extract their depth from the ground truth. On average, 400 points are extracted per image.

\begin{figure}[htbp]
  \centering
  \includegraphics[width=0.48\textwidth]{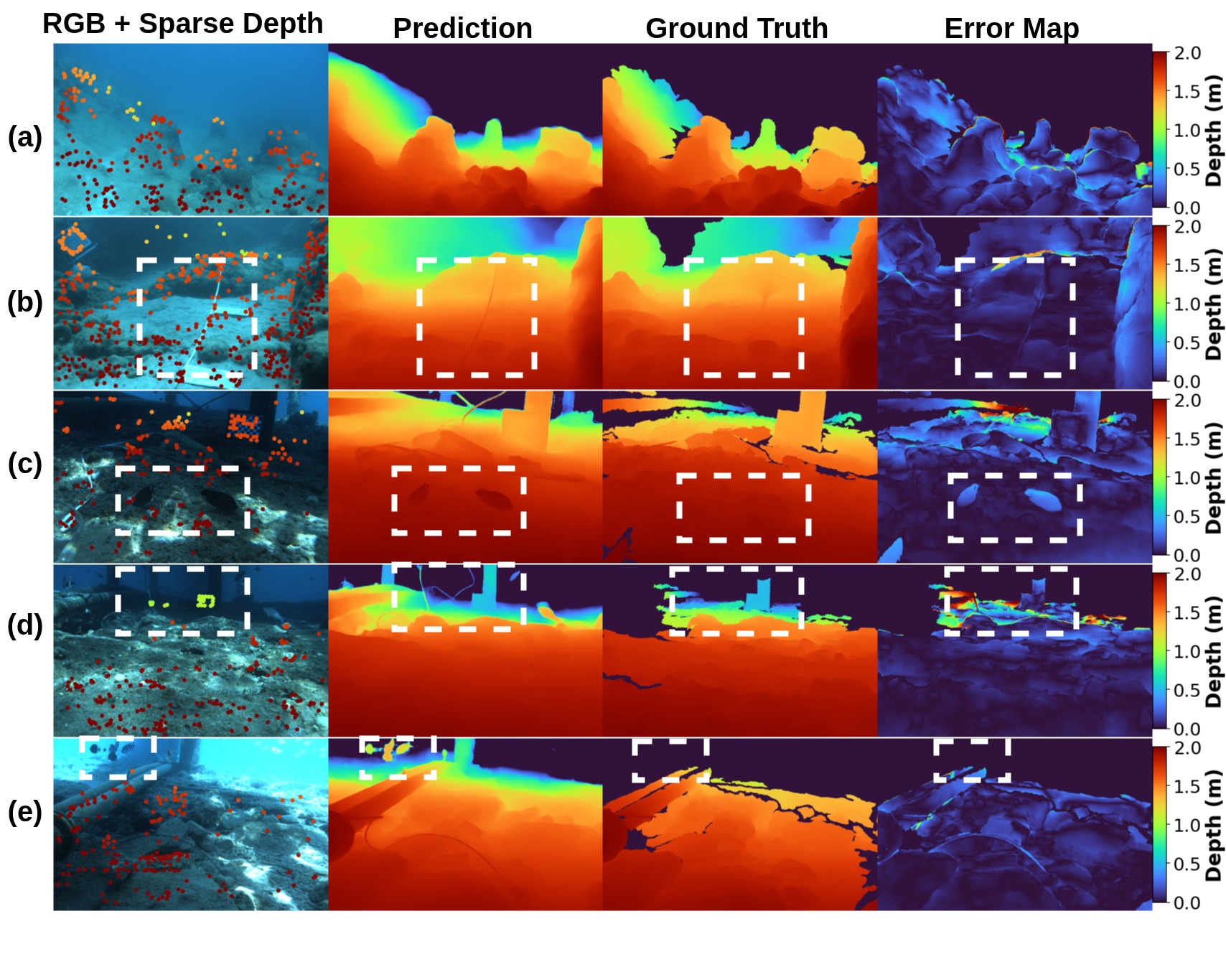} 
  \vspace{-10pt}
  \caption{\small Example results on the FLSea dataset. The boxes highlight dynamic or fine featured regions where the ground truth is missing data but the network predicts correctly.}
  \label{fig:flsea results few}
  \vspace{-10pt}
\end{figure}

Fig.~\ref{fig:flsea results few} presents qualitative results from our pipeline. These examples show that our method captures underwater depth with fine detail, including thin and dynamic objects such as tape measures, ropes, and fishes. However, these objects are often absent from the SfM ground truth, which is reflected in the error maps. The incomplete data in the SfM generated ground truth depth highlights the limitations of existing real-world underwater datasets for training and evaluation. 

Table~\ref{tab:flsea_range_result} shows quantitative results of our pipeline across different evaluation ranges. We limit the range to 10~m, as beyond this distance the visual signal is typically weak and noisy due to water attenuation and absorption. We compare against five recent monocular depth estimation methods that also use sparse depth measurements. These include approaches with and without a pretrained relative depth estimator. For a fair evaluation, we retrained most of the compared models on TartanAir, except Marigold-DC, a state-of-the-art depth completion method based on a relative depth diffusion model that was already trained on multiple dataset for generalisation performance. Note that, although the original VI-Depth model was already trained on the TartanAir dataset, it uses DPT-Hybrid \cite{midas3.1} as the relative depth estimator. We replaced that backbone with DepthAnythingV2 Small and retrained the model denoted as VI-Depth*. Table~\ref{tab:flsea_range_result} also shows that despite training solely on synthetic data, our model achieves the best accuracy across close, mid, and far range evaluations, consistently outperforming methods both with and without pretrained relative depth backbones on every metric, especially at longer evaluation range. Notably, while Marigold-DC achieves second best on some metrics and approaches our method's performance, its iterative diffusion makes it require over 14 seconds per image with 50 iterations (balancing speed and accuracy) on a desktop GPU. 
\begin{table*}[htbp]
  \centering
  \caption{\small Performance comparison of different methods on the FLSea dataset across varying evaluation ranges. This table reports the average metrics across 3270 frames of input, using an average of 400 prior depth points per frame (mean depth 2.37 m, median 2.02 m, max 6.75 m, min 1.21 m).}
  \scriptsize
  \setlength{\tabcolsep}{3pt}      % tighten horizontal padding
  \renewcommand{\arraystretch}{0.9} % tighten vertical padding
  \begin{tabular}{l c|cccc|cccc|cccc}
    \toprule
    \multicolumn{2}{c|}{\textbf{Range}} 
      & \multicolumn{4}{c|}{\textbf{$\leq$10\,m}}
      & \multicolumn{4}{c|}{\textbf{$\leq$5\,m}}
      & \multicolumn{4}{c}{\textbf{$\leq$2\,m}} \\
    \cmidrule(lr){1-2} \cmidrule(lr){3-6} \cmidrule(lr){7-10} \cmidrule(lr){11-14}
    \textbf{Method} & \textbf{Depth Backbone} 
      & MAE    & RMSE   & AbsRel & SILog 
      & MAE    & RMSE        & AbsRel & SILog 
      & MAE    & RMSE        & AbsRel & SILog \\
    \midrule
    UW-Depth~\cite{uwdepth}                   & No  & 0.174 & 0.393  & 0.058 & 0.097
                                       & 0.126 & \underline{0.236} & 0.054 & 0.081
                                       & 0.077 & 0.128 & 0.052 & 0.067 \\
    CompletionFormer~\cite{completionformer}            & No  & 0.227 & 0.581 & 0.133 & 0.213
                                       & 0.177 & 0.471 & 0.138 & 0.209
                                       & 0.335 & 0.731 & 0.686 & 0.332 \\
    DepthPrompting~\cite{DepthPrompting}              & Yes & 0.322 & 0.618 & 0.176 & 0.247
                                       & 0.272 & 0.496 & 0.178 & 0.247
                                       & 0.352 & 0.630 & 0.614 & 0.352 \\
    DA V2~\cite{DepthAnything}+GA            & Yes & 0.277 & 0.563 & 0.081 & 0.117
                                       & 0.170 & 0.296 & 0.068 & 0.095
                                       & 0.099 & 0.151 & 0.065 & 0.076 \\
    VI-Depth (Original)~\cite{videpth}       & Yes & 0.199 &0.561 &0.057 & 0.097
                                    &0.126 &0.319 &0.048 &0.076
                                    &0.063 &0.136 &0.041 & \textbf{0.054} \\ 
    VI-Depth*~\cite{videpth}                    & Yes & 0.180 & 0.398 & 0.053 & 0.090
                                       & 0.117 & 0.237   & 0.046 & 0.074
                                       & 0.067 & \underline{0.119} & 0.044 & 0.059 \\
    Marigold‑DC~\cite{MarigoldDC}                 & Yes & \underline{0.139} & 0.378 & \underline{0.044} & 0.095
                                       & \textbf{0.096}  & 0.247  & \underline{0.040} & 0.080
                                       & 0.062 & 0.145  & 0.042 & 0.068 \\
    \midrule
    Ours (PVT)       & Yes & 0.141 & \underline{0.354} & 0.045 & \underline{0.080}
                                       & \underline{0.101} & 0.247 & 0.041 & \underline{0.071}
                                       & \underline{0.056} & 0.121 & \underline{0.038} & \underline{0.055} \\
    Ours (Swin)      & Yes & 0.166 & 0.385 & 0.052 & 0.084
                                       & 0.118 & 0.263 & 0.047 & 0.074
                                       & 0.065 & 0.125 & 0.043 & 0.057 \\
    Ours (DAT)       & Yes & \textbf{0.131} & \textbf{0.316} & \textbf{0.042} & \textbf{0.077}
                                       & \textbf{0.096}    & \textbf{0.219} & \textbf{0.039} & \textbf{0.068}
                                       & \textbf{0.055}    & \textbf{0.115} & \textbf{0.037} & \textbf{0.054} \\
    \bottomrule
  \end{tabular}
  
  \label{tab:flsea_range_result}
  \vspace{-10pt}
\end{table*}

\begin{figure*}[htbp]
  \centering
  \includegraphics[width=0.7\textwidth]{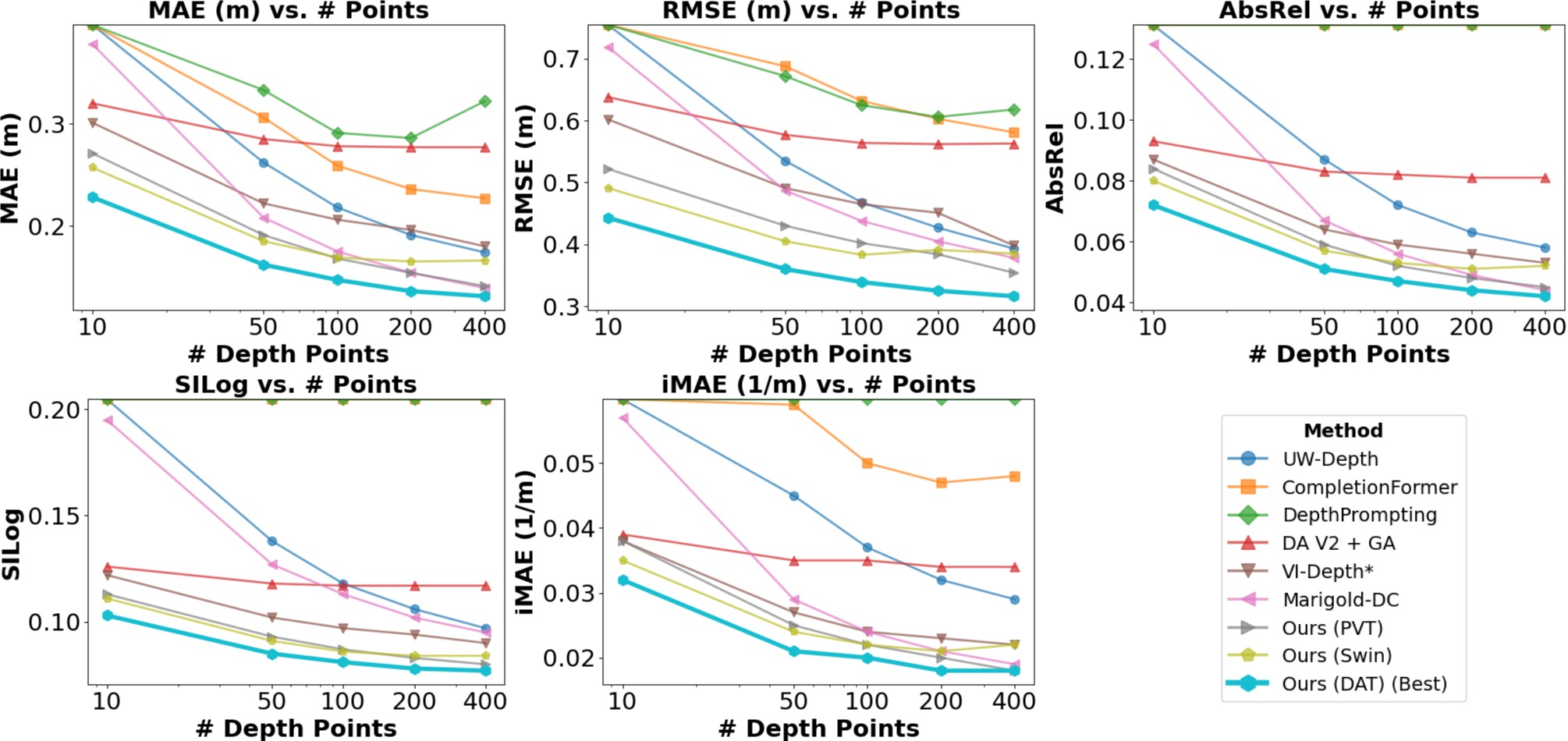} 
  \caption{\small Performance of methods evaluated in this study against different number of sparse depth points.}
  \label{fig:sparsity plot}
  \vspace{-15pt}
\end{figure*}

To assess robustness of our method to variations in depth points sparsity, we randomly subsampled the sparse depth points down to 200, 100, 50, and 10 per image and use these same subsets to evaluate all methods. As shown as Fig.~\ref{fig:sparsity_prediction}, our pipeline can predict reasonable depth maps with very sparse depth points. Fig.~\ref{fig:sparsity plot} quantifies how error increases for all methods as the number of sparse depth points decreases. To access the performance in close range, we also calculated MAE in inverse depth space denoted as iMAE. Our proposed method consistently achieves the lowest error across all metrics, with the performance gap becoming especially noticeable when very few sparse depth points are available. 

\begin{figure}[htbp]
  \centering
  \includegraphics[width=0.48\textwidth]{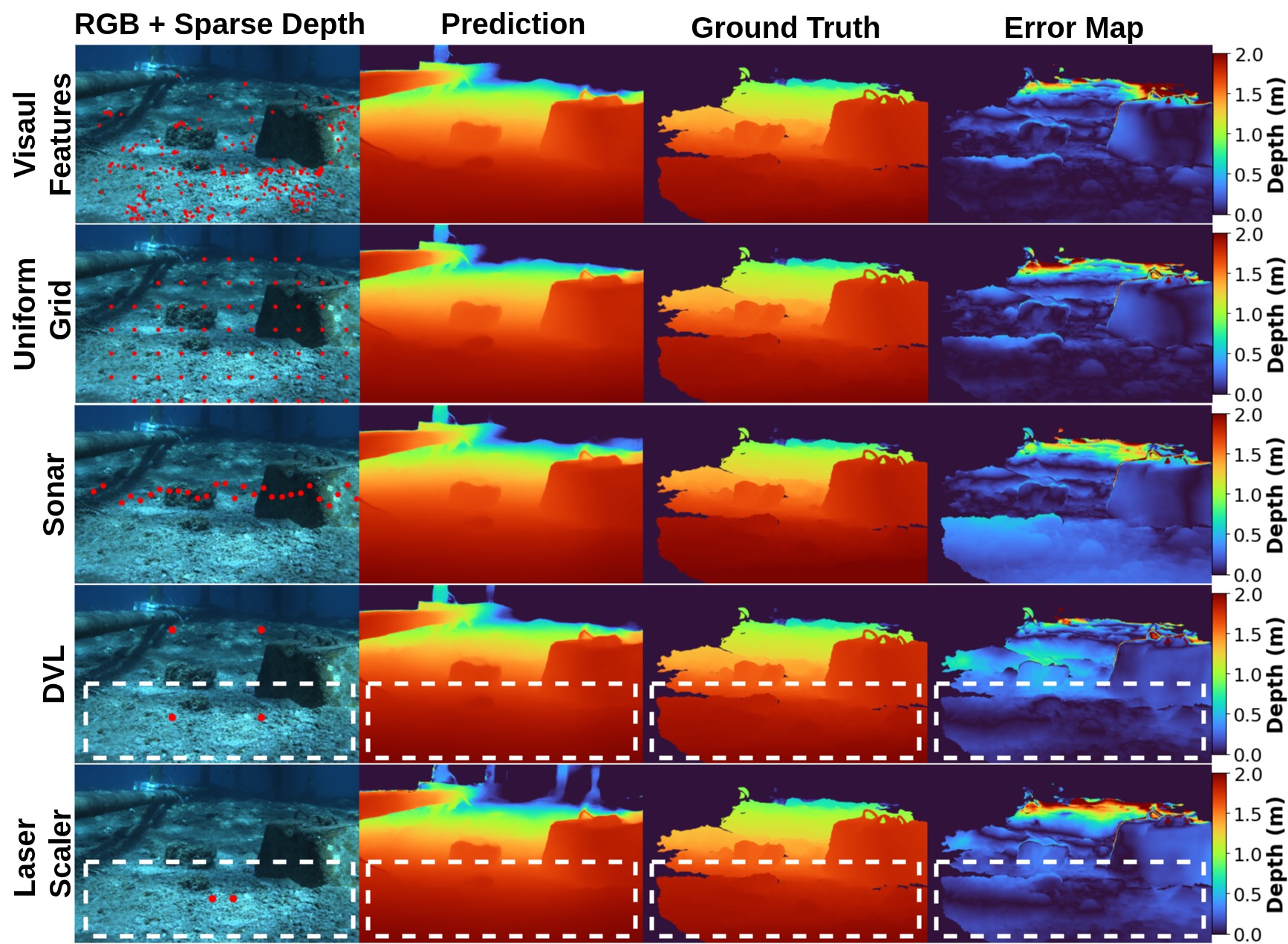} 
  \caption{\small Example results of different sparse depth distributions mimicking various depth sensors.}
  % \vspace{-10pt}
  \label{fig:distribution example}
  \vspace{-15pt}
\end{figure}

In our training and earlier evaluations, sparse depth points follow the scene’s visual feature distribution. We further assess our model’s robustness to different spatial distribution that might arise from alternative underwater sensing modalities, illustrated in Fig.~\ref{fig:distribution example}. The depth for these sparse points are sampled from the ground truth:
{
\begin{itemize}
   \item \textbf{Uniformly distribution sparse depth:} A uniformly distributed depth grid to simulate projected 3D sonar depth points or a fixed pattern laser projection.
  \item \textbf{Sonar/line-scanner:} A horizontal line of sparse depth points with random vertical offsets, simulating projections from a 2D imaging or scanning sonar
  \item \textbf{Doppler Velocity Logger (DVL):} Four isolated points located around the centre of the image to simulate a forward looking DVL providing 4 depth point reading.
  \item \textbf{Laser scaler:} Projection of two parallel lasers with 10cm baseline to simulate a laser scaler. As such laser scale cannot travel very long distance underwater, we only keep the projection within 3 metres. 
\end{itemize}
}
The global alignment step previously outlined requires at least three sparse points that are not aligned on the same depth plane to estimate a robust scale and shift. In the simulated laser scaler case with only two sparse points, we found a more robust approach to global alignment is to leverage the fixed baseline between the projected laser points to compute a global scale correction factor to the relative depth: 
\begin{equation}
  s_{laser} = \frac{\tfrac{u_2 - c_x}{f_x z_2} - \tfrac{u_1 - c_x}{f_x z_1}}{B},
\end{equation}
where $u_i$ is the pixel location of the $i^{\text{th}}$ projected laser point, $c_x$ is the camera optical centre, $f_x$ the camera focal length, $z_i$ is the relative depth at the $i^{\text{th}}$ projected laser point, and $B$ is the laser point baseline. 

As shown in Fig.~\ref{fig:distribution example}, our pipeline preserves accurate scene depth relationships and maintains low error across all sparse input patterns even with just two sparse depth points. While regions lacking sparse points have higher errors, areas that are geometrically or visually similar to known points still achieve reliable predictions. For instance, in the simulated DVL and laser scaler scenarios, the seabed at the bottom is predicted accurately despite most of the region being distant from the known depth points. Table~\ref{tab:flsea_distribution_result} further quantifies our method’s performance under different sparse depth distributions, demonstrating our pipeline's robustness. This evaluation also highlights that uniformly distributed sparse points yield more accurate depth maps than larger numbers of clustered visual feature points.
\begin{table*}[htbp]
  \centering
  \caption{\small Performance comparison of different methods on the FLSea dataset with varying sparse‐depth distributions (range 0–10 m).}
  \scriptsize
  \setlength{\tabcolsep}{2pt}
  \renewcommand{\arraystretch}{0.9}
  \begin{tabular}{l|ccccc|ccccc|ccccc|ccccc}
    \toprule
    \textbf{Distribution}
      & \multicolumn{5}{c|}{\textbf{Uniform Grid}}
      & \multicolumn{5}{c|}{\textbf{Simulated Sonar}}
      & \multicolumn{5}{c|}{\textbf{DVL (4 Points)}}
      & \multicolumn{5}{c}{\textbf{Laser Scaler (2 Points)}} \\
    \cmidrule(lr){1-1}\cmidrule(lr){2-6}\cmidrule(lr){7-11}\cmidrule(lr){12-16}\cmidrule(lr){17-21}
     \textbf{Method}
      & MAE    & RMSE  & AbsRel & SILog   & iMAE
      & MAE    & RMSE  & AbsRel & SILog   & iMAE
      & MAE    & RMSE  & AbsRel & SILog   & iMAE
      & MAE    & RMSE  & AbsRel & SILog   & iMAE \\
    \midrule
    UW-Depth~\cite{uwdepth}
      & 0.151 & \underline{0.324} & 0.054 & 0.086 & 0.027
      & 0.494 & 0.839 & 0.179 & 0.250 & 0.087
      & 0.403 & 0.677 & 0.158 & 0.211 & 0.085
      & 0.874 & 1.279 & 0.326 & 0.362 & 0.165 \\
    CompletionFormer~\cite{completionformer}
      & 0.175 & 0.514 & 0.119 & 0.200 & 0.038
      & 0.592 & 1.040 & 0.252 & 0.326 & 0.129
      & 0.621 & 0.958 & 0.279 & 0.305 & 0.169
      & 0.943 & 1.351 & 0.384 & 0.384 & 0.273 \\
    DepthPrompting~\cite{DepthPrompting}
      & 0.176 & 0.485 & 0.120 & 0.199 & 0.041
      & 0.536 & 0.901 & 0.244 & 0.318 & 0.139
      & 0.927 & 1.277 & 0.497 & 0.340 & 0.168
      & 1.061 & 1.397 & 0.566 & 0.345 & 0.195 \\
    DA V2~\cite{DepthAnything}+ GA
      & 0.242 & 0.459 & 0.077 & 0.110 & 0.033
      & 0.360 & 0.746 & 0.115 & 0.143 & \underline{0.048}
      & 0.247 & 0.494 & 0.086 & 0.114 & \underline{0.040}
      & 0.529 & 0.985 & 0.171 & 0.178 & \underline{0.065} \\
    VI-Depth*~\cite{videpth}
      & 0.170 & 0.398 & 0.053 & 0.087 & 0.022
      & 0.320 & 0.576 & 0.107 & 0.137 & 0.051
      & 0.246 & \underline{0.449} & 0.087 & 0.115 & 0.045
      & –     & –     & –     & –     & –     \\
    Marigold‑DC~\cite{MarigoldDC}
      & 0.135 & 0.365 & 0.048 & 0.096 & 0.020
      & 0.537 & 0.879 & 0.223 & 0.262 & 0.093
      & 0.474 & 0.767 & 0.205 & 0.248 & 0.092
      & 0.915 & 1.373 & 0.325 & 0.418 & 0.187 \\
    \midrule
    Ours (PVT)
      & \underline{0.131} & 0.329 & \underline{0.044} & \underline{0.075} & \underline{0.019}
      & 0.291 & \underline{0.556} & \underline{0.099} & \underline{0.129} & \underline{0.048}
      & \underline{0.236} & 0.450 & \underline{0.084} & \underline{0.111} & 0.042
      & 0.492 & 0.898 & 0.162 & 0.175 & 0.067 \\
    Ours (Swin)
      & 0.142 & 0.341 & 0.046 & 0.077 & \underline{0.019}
      & \underline{0.288} & \underline{0.556} & \underline{0.099} & 0.130 & \textbf{0.044}
      & 0.243 & 0.473 & 0.085 & 0.114 & \underline{0.040}
      & \underline{0.478} & \underline{0.891} & \underline{0.159} & \underline{0.174} & 0.067 \\
    Ours (DAT)
      & \textbf{0.116} & \textbf{0.282} & \textbf{0.039} & \textbf{0.070} & \textbf{0.017}
      & \textbf{0.259} & \textbf{0.481} & \textbf{0.090} & \textbf{0.119} & \textbf{0.044}
      & \textbf{0.213} & \textbf{0.398} & \textbf{0.076} & \textbf{0.102} & \textbf{0.038}
      & \textbf{0.409} & \textbf{0.708} & \textbf{0.138} & \textbf{0.152} & \textbf{0.062} \\
    \bottomrule
  \end{tabular}
  \vspace{-10pt}
  \label{tab:flsea_distribution_result}
\end{table*}

\subsection{Test with Lizard Island Dataset}
\begin{figure}[bp]
  \centering
  \includegraphics[width=0.48\textwidth]{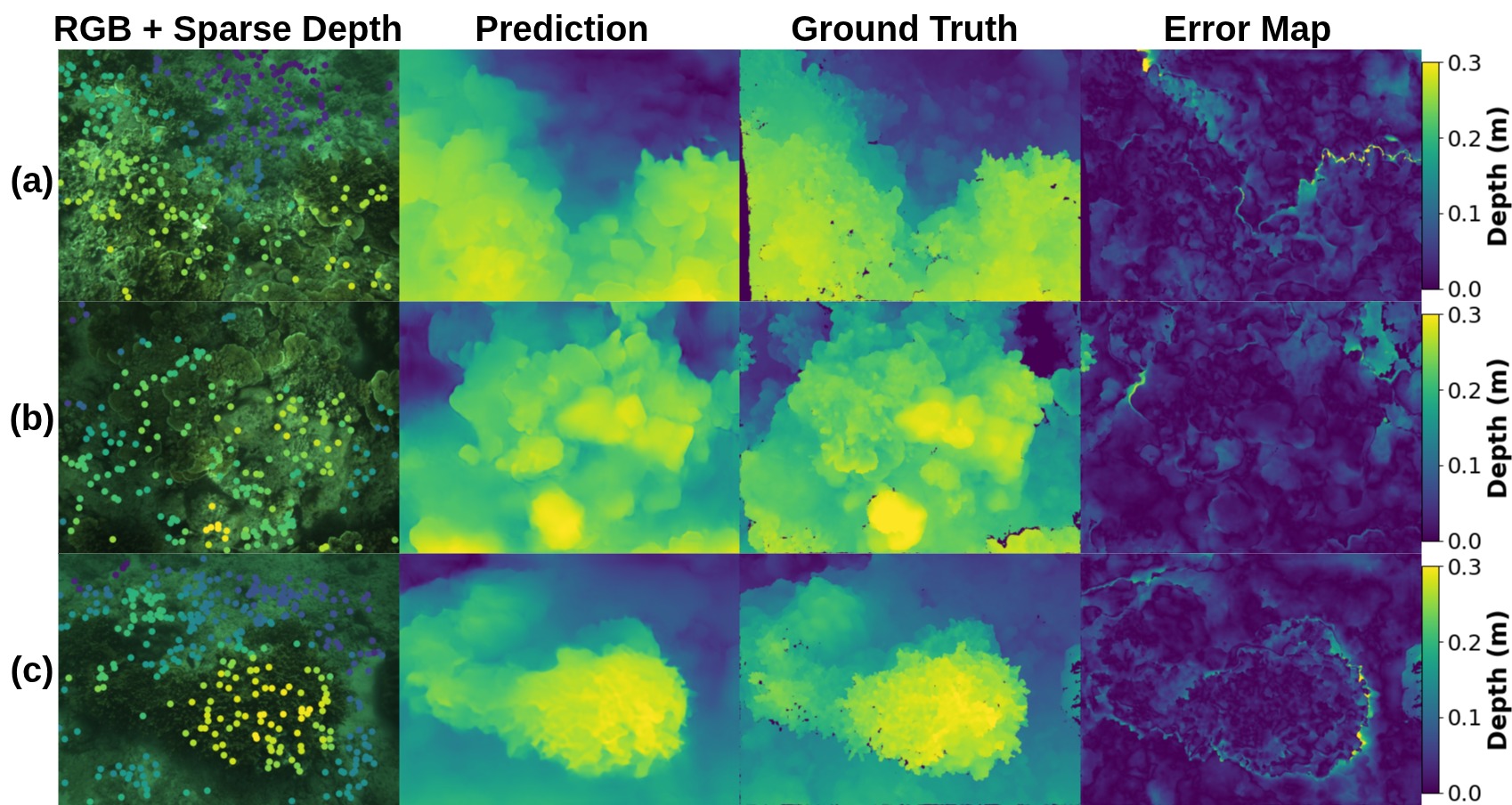} 
  \caption{\small Example results from the Lizard Island dataset.}
  \vspace{-10pt}
  \label{fig:lizard images}
\end{figure}

To assess our model’s ability to capture fine detail depth, we evaluated our model on the Lizard Island dataset \cite{CUDASiftSLAM}, which is a close-range, downward looking underwater scenario. This testing set comprises approximately 2,000 images of seafloor corals, where depth corresponds to the height difference between coral and the seabed. Ground truth depth maps are generated using COLMAP~\cite{COLMAP} from the stereo image sequence, and the sparse depth points are generated using ORB-SLAM2 adapted to SIFT feature points \cite{CUDASiftSLAM}. Fig.~\ref{fig:lizard images} presents qualitative results, which shows that our pipeline accurately captures the height differences of the coral, with most errors only occurring along the coral edges. We selected a colormap that changes monotonically to highlight detailed depth differences. Table~\ref{tab:lizard_result} reports the quantitative results. As this is a close range scenario, all methods exhibit relatively low prediction errors, however, our model still outperforms the others across every evaluation metric. 

\begin{table}[bp]
\centering
\caption{\small Performance comparison of different methods on the Lizard Island dataset. The performance is averaged over 2216 image frames with an average of 274 prior depth points per frame (mean depth: 1.93 m, median: 1.86 m, max: 3.53 m, min: 1.49 m) }
\label{tab:lizard_result}
\large
\begin{adjustbox}{max width=\columnwidth}
\begin{tabular}{lcccccc}
\hline
\textbf{Range} & \textbf{Method} & \textbf{AbsRel$\downarrow$} & \textbf{RMSE(m)$\downarrow$} & \textbf{MAE(m)$\downarrow$} & \textbf{SILog$\downarrow$}\\
\hline
\multirow{3}{*}{\textless 5 m} 
  & UW-Depth \cite{uwdepth}                    & 0.100 & 0.179  & 0.135 & 0.122\\
  & CompletionFormer \cite{completionformer}   & 0.064 & 0.125  & 0.059 & 0.137 \\
  & DA V2 \cite{DepthAnything} + GA     & 0.040 & 0.071 &0.054 &  0.050  \\
  & VI Depth* \cite{videpth}                  &\underline{0.031} & \underline{0.057} & \underline{0.042}  & \underline{0.039} \\
  & DepthPrompting \cite{DepthPrompting}      & 0.108 & 0.163  & 0.115 & 0.158  \\
  & Marigold-DC \cite{MarigoldDC}             & 0.035 & 0.096  & 0.046 & 0.054  \\
  & Ours (DAT)            & \textbf{0.025} & \textbf{0.051} & \textbf{0.034} & \textbf{0.034} \\
\hline
\end{tabular}
\end{adjustbox}
\vspace{-12pt}
\end{table}

\subsection{Water Tank Experiment}
Intervention tasks often require manipulation on delicate features, such as ropes, valves, or tools. However, capturing real-world datasets of such objects with accurate, dense ground truth is challenging, as SfM often reconstructs such features poorly. As a proof-of-concept for a future large-scale dataset, we collected a small dataset in a controlled water tank setting comprised of a metal frame with strung ropes of various thickness that could be accurately modelled in CAD. The dataset was collected using a BlueROV2 with a custom short-baseline stereo camera setup. Images were collected from multiple viewpoints and distances, and AprilTags were employed for precise pose recovery of the rig relative to the cameras. Dense ground truth depth was then rendered from a digital model of the rig using a virtual camera with parameters calibrated to the physical camera system. Sparse depth points were generated via single-shot stereo matching on the real images, with points on the AprilTags removed to reflect real-world conditions.

\begin{figure*}[tbp]
  \centering
  \includegraphics[width=0.95\textwidth]{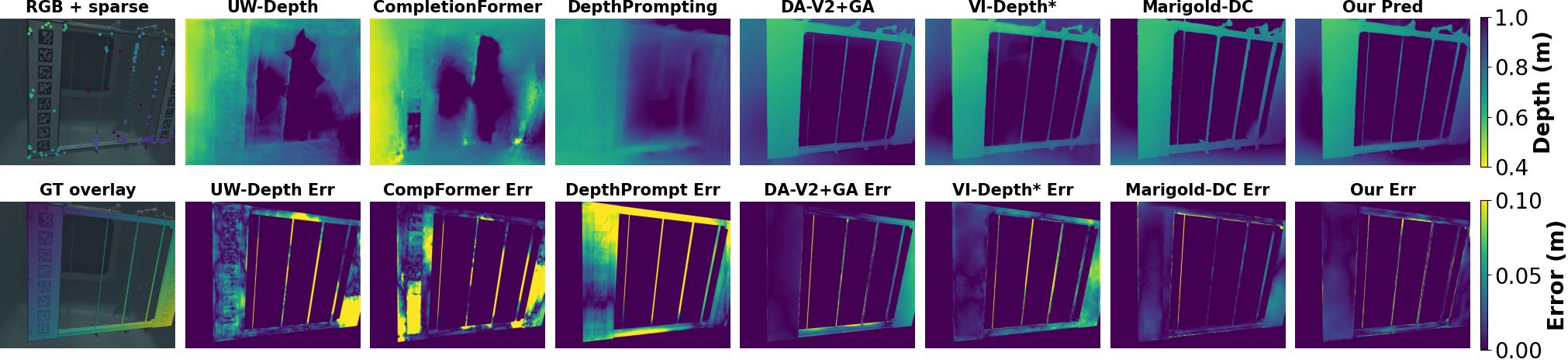} 
  \vspace{-5pt}
  \caption{\small Example results from the water tank experiment with real images and synthetically rendered depth.}% The CBAM block and the deformable attention block can have multiple layers by design.}
  \vspace{-10pt}
  \label{fig:rope object result}
\end{figure*}

Fig.~\ref{fig:rope object result} presents an image sample from this dataset and predictions from different methods. Depth points from single-shot stereo matching are very sparse in the background, given the low texture surface, and the thin ropes are only a few pixels wide on the subsampled images, making this a challenging scenario. Despite these challenges, our model achieves low depth error, while methods without a pretrained relative depth backbone perform poorly. We conjecture these methods suffer from a lack of close-range examples in the training data. Our method's predictions are noisier in the background than on the target, likely because of the homogeneous texture and lack of features in the background. For Marigold-DC, we used 10 iterations to balance overall quality and target error; more iterations reduced object error but introduced random background noise. As shown in Fig.~\ref{fig:rope object result}, Marigold-DC also achieves low error on the frame and ropes.

Table~\ref{tab:tank_rope_result} breaks down a detailed evaluation by rope thickness. As expected, performance degrades with decreasing thickness. Our model achieves the lowest errors on the 10 and 9~mm ropes. Marigold-DC \cite{MarigoldDC} slightly outperforms our method on the 6 mm rope and more noticeably on the 3~mm rope, but requires over 10 seconds to process each image. 
\begin{table}[tbp]
  \centering
  \caption{\small Evaluation of depth prediction on ropes of varying thickness in the water tank experiment. Evaluations used 34 frames of the same target within a 1.5~m range. Sparse depth points were obtained from single shot stereo matching with SIFT features. Features on AprilTags were removed.}
  \scriptsize
  \setlength{\tabcolsep}{2pt}      % tighten horizontal padding
  \renewcommand{\arraystretch}{0.9} % tighten vertical padding

  \begin{tabular}{l|cc|cc|cc|cc}
    \toprule
    \textbf{Thickness}
      & \multicolumn{2}{c|}{\textbf{10\,mm}}
      & \multicolumn{2}{c|}{\textbf{9\,mm}}
      & \multicolumn{2}{c|}{\textbf{6\,mm}}
      & \multicolumn{2}{c}{\textbf{3\,mm}} \\
    \cmidrule(lr){1-1}\cmidrule(lr){2-3}  \cmidrule(lr){4-5}  \cmidrule(lr){6-7}  \cmidrule(lr){8-9}
    \textbf{Method}
      & AbsRel  & MAE 
      & AbsRel  & MAE 
      & AbsRel  & MAE 
      & AbsRel  & MAE \\
    \midrule
    UW-Depth\cite{uwdepth}              & 0.097 & 0.082  & 0.123  & 0.102  & 0.118  & 0.099  & 0.101  & 0.083  \\
    CompletionFormer\cite{completionformer}      & 0.713 & 0.219  & 1.008  & 0.351  & 1.046  & 0.298  & 1.284  & 0.323  \\
    DepthPrompting\cite{DepthPrompting}        & 0.895  & 0.244  & 1.041  & 0.291  & 1.372  & 0.369  & 1.875  & 0.471  \\
    DA V2+GA\cite{DepthAnything}      & 0.060  & 0.057  & 0.060  & 0.054  & 0.096  & 0.081  & 0.158  & 0.127  \\
    VI-Depth*\cite{videpth}             & 0.075  & 0.068  & 0.066  & 0.059  & 0.068  & 0.060  & 0.094  & 0.079  \\
    Marigold‑DC\cite{MarigoldDC}           & \underline{0.052}  & \underline{0.047}
                          & 0.066           & 0.057
                          & \textbf{0.062}     & \textbf{0.053}
                          & \textbf{0.059}      & \textbf{0.049} \\
    Ours (DAT)       & \textbf{0.044}    & \textbf{0.038}
                          & \textbf{0.050}       & \textbf{0.043}
                          & \textbf{0.062}     & \underline{0.054}
                          & \underline{0.085}  & \underline{0.074} \\
    \bottomrule
  \end{tabular}

  \vspace{-6pt}
  \label{tab:tank_rope_result}
  \vspace{-6pt}
\end{table}

\section{Discussion}
Zero-shot evaluations demonstrate that our proposed monocular depth estimation pipeline reliably predicts dense metric depth maps across diverse underwater scenes and camera settings, without special fine-tuning. This strong generalisation arises from our approach of predicting a scale correction map to convert the arbitrary-scale depth predictions of a relative depth estimator into metric scale. By leveraging a relative depth estimator that was pre-trained on large-scale, diverse datasets, our method inherits its robust scene geometry generalisation. Predicting a scale correction map, rather than directly predicting metric depth values, also reduces the risk of overfitting to specific depth ranges. This approach reduces reliance on large-scale, high-quality underwater training data for achieving generalisation performance. Similar strategies have been adopted in \cite{videpth} and \cite{MarigoldDC}, which also demonstrate strong generalisability across different underwater scenes in our evaluations. In contrast, methods that do not use a relative backbone and directly predict metric depth maps, such as \cite{uwdepth} and \cite{completionformer}, tend to degrade in performance more on images outside of the training distribution.

The choice of DepthAnythingV2 as our relative depth backbone contributes to improved accuracy and consistency across diverse scene ranges. While the original VI-Depth implementation~\cite{videpth} achieved high performance at close range, the retrained model with DepthAnythingV2 achieves better performance at greater ranges, highlighting the advantages of a well-generalised relative depth estimator. In contrast, DepthPrompting~\cite{DepthPrompting}, which uses a relative depth backbone trained on a single dataset, lacks this generalisation and exhibits higher errors in underwater tests. 

Despite its strengths, the small variant of DepthAnythingV2 has two limitations. First, unique underwater lighting effects, such as caustics, can reduce relative depth prediction consistency in regions with rapidly changing illumination. This could be mitigated by pre-processing for caustic removal or fine-tuning the model on representative underwater images while preserving generalisation. Second, the small variant sacrifices fine detail for speed. Although the mean depth prediction is very good, it struggles with very thin objects, such as the 3 mm rope in our water tank dataset, compared to Marigold-DC, which uses a larger relative depth backbone. An avenue worth exploring is whether larger depth backbones with quantisation could boost performance on fine structures while retaining real-time inference on embedded compute. In addition, it would be valuable to investigate the trade-offs between model size, processed image resolution, reconstruction detail, and computational efficiency.

To quantify the improvement provided by our scale refinement network, we compared the intermediate output of DepthAnythingV2 small with global alignment (DA V2 Small+GA) to the final output of our method. Our full pipeline consistently outperforms the intermediate results across all benchmarks. As shown in the sparsity analysis (Fig. \ref{fig:sparsity plot}), the performance of DA V2 Small+GA saturates as more points are added, while our per-pixel scale correction strategy continues to reduce error. This highlights that the scale refinement network is capable of leveraging the known sparse depth points to correct potential geometric biases from the relative depth estimator.

Our evaluations demonstrate that our proposed pipeline generally outperforms alternatives with both many and few sparse depth points and across varying spatial distribution. This robustness arises from our proposed cascade convolutional and transformer block in the refinement network. Our pipeline generally outperforms the re-trained VI-Depth, which uses the EfficientNet CNN encoder for scale correction, although both models using the same relative-depth backbone and two-stage design. By employing our proposed block in the refinement network, we achieve lower errors with a comparable number of parameters. Furthermore, our method consistently outperforms other methods across the varying sparsity and distribution tests (Fig. \ref{fig:sparsity plot} and Table~\ref{tab:flsea_distribution_result}). These results verify that our cascade block effectively combines the strengths of convolutional and transformer modules, allowing it to relate unknown regions to known sparse depth and scale points, both locally and over long distances, thereby demonstrating robustness to different depth points sparsity and spatial distributions.

To further assess the benefit of deformable attention in the proposed CCDT encoder block, we replaced the DAT module with Swin Transformer and Pyramid Vision Transformer (PVT) in an ablation study on the FLSea dataset. DAT consistently outperformed both alternatives. With dense depth points, DAT and PVT achieved similar results (Table~\ref{tab:flsea_range_result}); however, as sparsity increased and distributions became uneven (Fig.~\ref{fig:sparsity plot} and Table~\ref{tab:flsea_distribution_result}), DAT maintained superior accuracy. These results highlight the advantage of DAT’s adaptive attention patterns over the fixed schemes used by other efficient transformers.

Our evaluation of sparse depth distributions by simulating different sensor modalities also highlights our method’s flexibility of adapting to commonly used sensors (e.g., DVL, laser projector, imaging sonar) as sources of sparse depth points. Notably, the spatial distribution results (Table~\ref{tab:flsea_distribution_result}) show that a uniform depth grid produces lower errors than a dense cluster of visual features, even when more points are available in the latter. This underscores the advantage of integrating sensor modalities that provide uniformly distributed depth points, especially in texture-less regions, and highlights the value of multi-modal fusion for reliable underwater depth perception, which is a direction we plan to explore in future work.

\section{Conclusion}
We introduce a monocular depth estimation pipeline for real-time underwater applications. Through extensive evaluation, we show that our proposed method demonstrates improved performance over the state-of-the-art across multiple baseline methods. Our modular design lets users swap the sparse generation module with SLAM approaches or single‑shot stereo matching and upgrade the relative depth backbone for finer detail. Our method also supports straightforward integration of alternative depth sensors by projecting their measurements into the camera frame. In a lightweight, real‑time setup, the pipeline runs at over 15 FPS on an NVIDIA Jetson ORIN NX, making it suitable for navigation, obstacle avoidance, and underwater intervention tasks. Our evaluation with emulated sensor inputs highlights the advantages of integrating additional sensor modalities to improve accuracy and robustness feature poor environments. Future work will focus on multi-modal fusion of cameras with active sensors to further enhance underwater spatial awareness.

\bibliographystyle{IEEEtran}
\bibliography{root}

\begin{thebibliography}{10}
\providecommand{\url}[1]{#1}
\csname url@samestyle\endcsname
\providecommand{\newblock}{\relax}
\providecommand{\bibinfo}[2]{#2}
\providecommand{\BIBentrySTDinterwordspacing}{\spaceskip=0pt\relax}
\providecommand{\BIBentryALTinterwordstretchfactor}{4}
\providecommand{\BIBentryALTinterwordspacing}{\spaceskip=\fontdimen2\font plus
\BIBentryALTinterwordstretchfactor\fontdimen3\font minus \fontdimen4\font\relax}
\providecommand{\BIBforeignlanguage}[2]{{%
\expandafter\ifx\csname l@#1\endcsname\relax
\typeout{** WARNING: IEEEtran.bst: No hyphenation pattern has been}%
\typeout{** loaded for the language `#1'. Using the pattern for}%
\typeout{** the default language instead.}%
\else
\language=\csname l@#1\endcsname
\fi
#2}}
\providecommand{\BIBdecl}{\relax}
\BIBdecl

\bibitem{phung2024shared}
A.~Phung, G.~Billings, A.~F. Daniele, M.~R. Walter, and R.~Camilli, ``A shared autonomy system for precise and efficient remote underwater manipulation,'' \emph{IEEE Trans. Robot.}, vol.~40, pp. 4147--4159, 2024, doi: 10.1109/TRO.2024.3431830.

\bibitem{FLSea}
Y.~Randall and T.~Treibitz, ``Flsea: Underwater visual-inertial and stereo-vision forward-looking datasets,'' 2023, \textit{arXiv:2302.12772}.

\bibitem{mono_single_beam}
M.~Roznere and A.~Q. Li, ``Underwater monocular image depth estimation using single-beam echosounder,'' in \emph{Proc. 2020 IEEE/RSJ Int. Conf. Intell. Robots Syst. (IROS)}, Oct. 2020, pp. 1785--1790.

\bibitem{camera_sonar_slam}
A.~Cardaillac and M.~Ludvigsen, ``Camera-sonar combination for improved underwater localization and mapping,'' \emph{IEEE Access}, vol.~11, pp. 123\,070--123\,079, 2023, doi: 10.1109/ACCESS.2023.3329834.

\bibitem{midas3.1}
R.~Birkl, D.~Wofk, and M.~M{\"u}ller, ``Midas v3.1 -- a model zoo for robust monocular relative depth estimation,'' 2023, \textit{arXiv:2307.14460}.

\bibitem{DepthAnything}
L.~Yang, B.~Kang, Z.~Huang, Z.~Zhao, X.~Xu, J.~Feng \emph{et~al.}, ``Depth anything v2,'' 2024, \textit{arXiv:2406.09414}.

\bibitem{marigold}
B.~Ke, A.~Obukhov, S.~Huang, N.~Metzger, R.~C. Daudt, and K.~Schindler, ``Repurposing diffusion-based image generators for monocular depth estimation,'' in \emph{Proc. 2024 IEEE/CVF Conf. Comput. Vis. Pattern Recognit. (CVPR)}, Jun. 2024, pp. 9492--9502.

\bibitem{Atlantis}
F.~Zhang, S.~You, Y.~Li, and Y.~Fu, ``Atlantis: Enabling underwater depth estimation with stable diffusion,'' in \emph{Proc. 2024 IEEE/CVF Conf. Comput. Vis. Pattern Recognit. (CVPR)}, Jun. 2024, pp. 11\,852--11\,861.

\bibitem{videpth}
D.~Wofk, R.~Ranftl, M.~Müller, and V.~Koltun, ``Monocular visual-inertial depth estimation,'' in \emph{Proc. 2023 IEEE Int. Conf. Robot. Autom. (ICRA)}, May 2023, pp. 6095--6101.

\bibitem{completionformer}
Y.~Zhang, X.~Guo, M.~Poggi, Z.~Zhu, G.~Huang, and S.~Mattoccia, ``Completionformer: Depth completion with convolutions and vision transformers,'' in \emph{Proc. 2023 IEEE/CVF Conf. Comput. Vis. Pattern Recognit. (CVPR)}, Jun. 2023, pp. 18\,527--18\,536.

\bibitem{SparseDC}
C.~Long, W.~Zhang, Z.~Chen, H.~Wang, Y.~Liu, P.~Tong \emph{et~al.}, ``Sparsedc: Depth completion from sparse and non-uniform inputs,'' \emph{Inf. Fusion}, vol. 110, p. 102470, Oct. 2024, doi: 10.1016/j.inffus.2024.102470.

\bibitem{SpAgNet}
A.~Conti, M.~Poggi, and S.~Mattoccia, ``Sparsity agnostic depth completion,'' in \emph{Proc. 2023 IEEE/CVF Winter Conf. Appl. Comput. Vis. (WACV)}, Jan. 2023, pp. 5871--5880.

\bibitem{FlexibleDC}
J.~Park, Y.-J. Li, and K.~Kitani, ``Flexible depth completion for sparse and varying point densities,'' in \emph{Proc. 2024 IEEE/CVF Conf. Comput. Vis. Pattern Recognit. (CVPR)}, Jun. 2024, pp. 21\,540--21\,550.

\bibitem{NLSPN}
J.~Park, K.~Joo, Z.~Hu, C.-K. Liu, and I.~S. Kweon, ``Non-local spatial propagation network for depth completion,'' in \emph{Proc. 16th Eur. Conf. Comput. Vis. (ECCV)}, Aug. 2020, pp. 120--136.

\bibitem{VirtualStereo}
L.~Bartolomei, M.~Poggi, A.~Conti, F.~Tosi, and S.~Mattoccia, ``Revisiting depth completion from a stereo matching perspective for cross-domain generalization,'' in \emph{Proc. 2024 Int. Conf. 3D Vis. (3DV)}, Mar. 2024, pp. 1360--1370.

\bibitem{uwdepth}
L.~Ebner, G.~Billings, and S.~Williams, ``Metrically scaled monocular depth estimation through sparse priors for underwater robots,'' in \emph{Proc. 2024 IEEE Int. Conf. Robot. Autom. (ICRA)}, May 2024, pp. 3751--3757.

\bibitem{DepthPrompting}
J.-H. Park, C.~Jeong, J.~Lee, and H.-G. Jeon, ``Depth prompting for sensor-agnostic depth estimation,'' in \emph{Proc. 2024 IEEE/CVF Conf. Comput. Vis. Pattern Recognit. (CVPR)}, Jun. 2024, pp. 9859--9869.

\bibitem{MarigoldDC}
M.~Viola, K.~Qu, N.~Metzger, B.~Ke, A.~Becker, K.~Schindler \emph{et~al.}, ``Marigold-dc: Zero-shot monocular depth completion with guided diffusion,'' 2024, \textit{arXiv:2412.13389}.

\bibitem{CUDASiftSLAM}
G.~Billings, R.~Camilli, and M.~Johnson-Roberson, ``Hybrid visual slam for underwater vehicle manipulator systems,'' \emph{IEEE Robot. Autom. Lett.}, vol.~7, no.~3, pp. 6798--6805, Jul. 2022, doi: 10.1109/LRA.2022.3176448.

\bibitem{U-Net}
O.~Ronneberger, P.~Fischer, and T.~Brox, ``U-net: Convolutional networks for biomedical image segmentation,'' in \emph{Proc. 18th Int. Conf. Med. Image Comput. Comput.-Assist. Intervent. (MICCAI)}, Oct. 2015, pp. 234--241.

\bibitem{DPT}
R.~Ranftl, A.~Bochkovskiy, and V.~Koltun, ``Vision transformers for dense prediction,'' in \emph{Proc. 2021 IEEE/CVF Int. Conf. Comput. Vis. (ICCV)}, Oct. 2021, pp. 12\,179--12\,188.

\bibitem{joint_bilateral_upsampling}
J.~Kopf, M.~F. Cohen, D.~Lischinski, and M.~Uyttendaele, ``Joint bilateral upsampling,'' \emph{ACM Trans. Graph.}, vol.~26, no.~3, p.~96, Jul. 2007, doi: 10.1145/1276377.1276497.

\bibitem{CBAM}
S.~Woo, J.~Park, J.-Y. Lee, and I.~S. Kweon, ``Cbam: Convolutional block attention module,'' in \emph{Proc. 15th Eur. Conf. Comput. Vis. (ECCV)}, Sep. 2018, pp. 3--19.

\bibitem{dat++}
Z.~Xia, X.~Pan, S.~Song, L.~E. Li, and G.~Huang, ``Dat++: Spatially dynamic vision transformer with deformable attention,'' 2023, \textit{arXiv:2309.01430}.

\bibitem{vit}
A.~Dosovitskiy, L.~Beyer, A.~Kolesnikov, D.~Weissenborn, X.~Zhai, T.~Unterthiner \emph{et~al.}, ``An image is worth 16x16 words: Transformers for image recognition at scale,'' 2021, \textit{arXiv:2010.11929}.

\bibitem{Swin}
Z.~Liu, Y.~Lin, Y.~Cao, H.~Hu, Y.~Wei, Z.~Zhang \emph{et~al.}, ``Swin transformer: Hierarchical vision transformer using shifted windows,'' in \emph{Proc. 2021 IEEE/CVF Int. Conf. Comput. Vis. (ICCV)}, Oct. 2021, pp. 10\,012--10\,022.

\bibitem{PVT}
W.~Wang, E.~Xie, X.~Li, D.-P. Fan, K.~Song, D.~Liang \emph{et~al.}, ``Pyramid vision transformer: A versatile backbone for dense prediction without convolutions,'' in \emph{Proc. 2021 IEEE/CVF Int. Conf. Comput. Vis. (ICCV)}, Oct. 2021, pp. 548--558.

\bibitem{TartanAir}
W.~Wang, D.~Zhu, X.~Wang, Y.~Hu, Y.~Qiu, C.~Wang \emph{et~al.}, ``Tartanair: A dataset to push the limits of visual slam,'' in \emph{Proc. 2020 IEEE/RSJ Int. Conf. Intell. Robots Syst. (IROS)}, Oct. 2020, pp. 4909--4916.

\bibitem{VINS-MONO}
T.~Qin, P.~Li, and S.~Shen, ``Vins-mono: A robust and versatile monocular visual-inertial state estimator,'' \emph{IEEE Trans. Robot.}, vol.~34, no.~4, pp. 1004--1020, Aug. 2018, doi: 10.1109/TRO.2018.2853729.

\bibitem{COLMAP}
J.~L. Schonberger and J.-M. Frahm, ``Structure-from-motion revisited,'' in \emph{Proc. 2016 IEEE/CVF Conf. Comput. Vis. Pattern Recognit. (CVPR)}, Jun. 2016, pp. 4104--4113.

\end{thebibliography}

\end{document}